\documentclass{article}
\usepackage{natbib}
\usepackage{latexsym}
\usepackage{amssymb}
\usepackage{amsmath}
\usepackage{amsthm}
\usepackage{booktabs}
\usepackage{enumitem}
\usepackage{graphicx}
\usepackage{times}
\usepackage{soul}
\usepackage{url}
\usepackage[hidelinks]{hyperref}
\usepackage[utf8]{inputenc}
\usepackage[small]{caption}
\usepackage{graphicx}
\usepackage{amsmath}
\usepackage{amsthm}
\usepackage{booktabs}
\usepackage{algorithm}
\usepackage{algorithmic}
\usepackage{booktabs}
\usepackage{amsmath}  
\usepackage{amssymb}  
\usepackage{amsfonts} 
\usepackage{comment}
\usepackage{amssymb}
\usepackage{arxiv}
\usepackage{xcolor}

\title{P-CAFE: Personalized Cost-Aware Incremental Feature Selection For Electronic Health Records}

\author{
	Naama Kashani\thanks{Equal contribution, random order.} \\
	Department of Computer Science\\
	Bar-Ilan University\\
	Ramat-Gan, Israel \\
	\texttt{naamakashani@gmail.com} \\
    \And
    Mira Cohen\footnotemark[1]\\
	\texttt{miracohen7@gmail.com} \\
	\AND
	Uri Shaham \\
	Department of Computer Science\\
	Bar-Ilan University\\
	Ramat-Gan, Israel \\
	\texttt{uri.shaham@biu.ac.il} \\
}

\date{}

\begin{document}
\maketitle

\begin{abstract}
Electronic Health Records (EHR) have revolutionized healthcare by digitizing patient data, improving accessibility, and streamlining clinical workflows. However, extracting meaningful insights from these complex and multimodal datasets remains a significant challenge for researchers. Traditional feature selection methods often struggle with the inherent sparsity and heterogeneity of EHR data, especially when accounting for patient-specific variations and feature costs in clinical applications. To address these challenges, we propose a novel personalized, online and cost-aware feature selection framework tailored specifically for EHR datasets. The features are aquired in an online fashion for individual patients, incorporating budgetary constraints and feature variability costs. The framework is designed to effectively manage sparse and multimodal data, ensuring robust and scalable performance in diverse healthcare contexts. A primary application of our proposed method is to support physicians' decision making in patient screening scenarios. By guiding physicians toward incremental acquisition of the most informative features within budget constraints, our approach aims to increase diagnostic confidence while optimizing resource utilization.

\end{abstract}

\section{Introduction}
Electronic Health Records (EHRs) serve as comprehensive digital repositories of patient health information, encompassing both structured and unstructured data \citep{bates2014big}. A thorough understanding of EHR data can significantly enhance various aspects of patient care, including disease prediction, healthcare quality improvement, and resource allocation \citep{8086133, KIM2019354}. However, EHR data presents unique challenges: it is often high-dimensional, multimodal, sparse, and temporal \citep{wu2010prediction, doi:10.2147/RMHP.S12985, 10.1093/jamia/ocy068}. Records typically include a diverse array of modalities, such as demographics, diagnoses, procedures, medications, prescriptions, radiological images, clinical notes, and laboratory results. The data is inherently sparse, as medical events occur irregularly, and sequential, as patient histories accumulate over time.

To address these complexities, many approaches employ feature selection (FS) — the process of identifying the most informative variables from high-dimensional input to improve model performance, interpretability, and robustness \citep{REMESEIRO2019103375, chandrashekar2014survey}. Yet, to the best of our knowledge, existing FS methods applied to EHRs either ignore multimodality or fail to capture temporal dynamics. Moreover, classical FS techniques assume that all features are available upfront and aim to identify a single subset applicable to all samples. In contrast, clinical decision-making is inherently dynamic and personalized: patient information is acquired incrementally, and medical tests vary greatly in cost. These aspects of EHRs underscore the need for FS methods that can mimic the real-world, stepwise, and cost-sensitive nature of clinical reasoning, while accounting for multimodality, temporal structure, sparsity, and heterogeneity in acquisition cost.

Inspired by the clinical reasoning process employed by healthcare professionals, we propose a new FS paradigm. During a patient consultation, a physician gathers information iteratively — asking targeted questions, performing physical examinations, and ordering diagnostic tests — with the goal of maximizing diagnostic value while minimizing unnecessary procedures. This process is personalized and context-aware: each step depends on previously acquired information. Conceptually, it can be viewed as a human-driven online FS procedure.

\begin{figure}[H]
    \centering
    \includegraphics[width=1\linewidth]{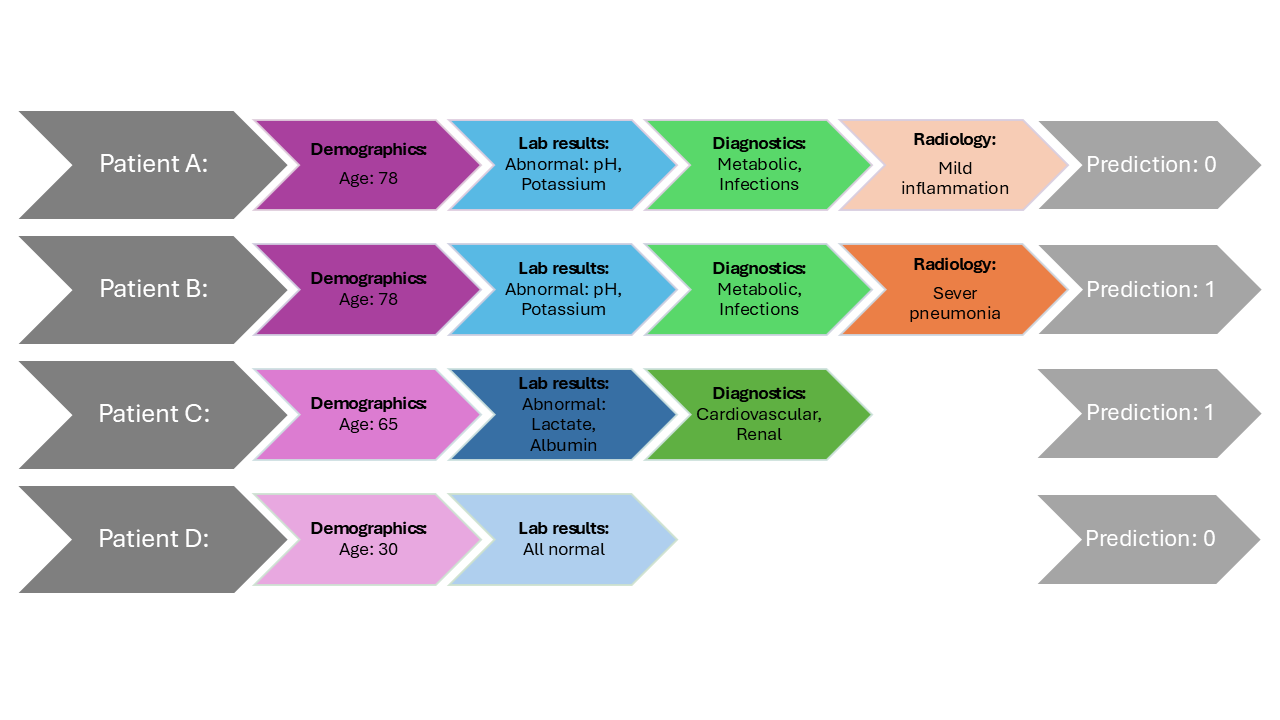} 
    \caption{P-CAFE framework applied to patient-specific cases of the MIMIC-III Multi-Modal Dataset. The progression through the feature selection stages is online and tailored to each patient, with predictions of in-hospital mortality displayed on the right to reflect personalized outcomes.}
    \label{fig:cases}
\end{figure}

Motivated by this insight, we introduce \textbf{P-CAFE}, a novel, comprehensive feature selection framework designed to address the unique challenges of EHR data. P-CAFE explicitly handles high-dimensional, multimodal inputs; captures the temporal structure of patient histories; accounts for data sparsity \citep{Getzen2022.05.09.22274680}; and models variable feature acquisition costs — all while emulating the personalized and incremental nature of clinical consultations. Crucially, P-CAFE is designed to optimize outcome prediction by selectively acquiring the most informative features for each patient. As illustrated in Figure~\ref{fig:cases}, P-CAFE performs sequential, patient-specific feature selection over multimodal EHR inputs, adapting its decisions based on the information accumulated thus far to maximize predictive utility while minimizing acquisition cost.

Ultimately, our approach empowers physicians by providing insights into both the expected benefit — e.g., improved diagnostic accuracy — and the associated cost of acquiring additional medical information. This enables more informed, efficient, and personalized decisions at each stage of the diagnostic process. P-CAFE is readily applicable as a clinical decision-support tool aimed at guiding cost-effective and patient-tailored diagnostic strategies.

\section{Related Work}
\textbf {Feature Selection On Electronic Health Records}
Previous research of FS on EHR datasets has focused on identifying a global feature subset; this lack of personalization in EHRs can lead to inaccurate treatment plans and missed diagnoses. The methods of \cite{bhadra2024enhancing}, \cite{zuo2021curvature}, and \cite{ebrahimi2022identification} use various global FS techniques and multiple machine learning models to predict the outcome of EHR data. Similar to \cite{tsang2020modeling}, our design incorporates feedback on the predictive model's performance based on the selected features. Entropy regularization with ensemble deep neural networks was utilized to perform global feature selection while training predictive models on EHR data, as shown in \cite{tsang2020modeling}. In \cite{dhinakaran2024optimizing}, a global feature selection method was integrated with a deep network and long short-term memory (LSTM) to process EHR data.

\textbf {Personalized Feature Selection} 
Existing personalized FS approaches, such as \cite{7265078}, \cite{shaham2020learning}, \cite{yang2022locally}, and \cite{shahrjooihaghighi2022local}, are specifically designed for tabular datasets, are often not designed to account for feature costs, and are not equipped to handle the challenges posed by sparse, multimodal and time-series data. As a result, they are not suited for EHRs, where these characteristics are prominent.

\textbf {Feature Selection Using Reinforcement Learning} 
The use of reinforcement learning (RL) for FS is particularly advantageous due to its ability to handle scenarios where no clear ``ground truth" or predefined correct feature subset exists. Instead, RL leverages rewards as a form of supervision, enabling the model to evaluate and adapt based on the performance of selected features, thereby guiding the feature selection process effectively. RL-based algorithms offer a flexible framework that allows exploration of different feature combinations, evaluation through rewards, and continuous adaptation to the data's characteristics.  Like other RL-based FS methods, we define the FS problem as a Markov Decision Process (MDP) and use an agent to select the optimal subset of features. However, our approach is distinct because it performs FS in a personalized manner, unlike other RL-based FS methods such as \cite{rasoul2021feature}, \cite{HAZRATIFARD20131892}, \cite{fan2020autofs},\cite{kim2022feature}, which apply a global FS strategy.

Recent developments in this area include the work by \cite{chen2024dynamic}, who proposes an FS method using RL for medical predictive monitoring in multivariate time-series scenarios, which is particularly relevant to EHR data analysis. Additionally, \cite{wu2023value} introduced a value-based deep RL model with human expertise for sepsis treatment, which incorporates FS to improve model interpretability. In contrast, our approach advances the application of FS for EHR data by incorporating personalized, cost-aware FS and enabling the integration of multimodal data types, factors that were not comprehensively considered.

\textbf {Feature Selection Via Masking}
Online FS using masking has been explored in studies such as \cite{turali2024afs} and \cite{lorasdagi2024binary}. These works introduced methods for adaptive feature selection through binary masking, enabling the simultaneous optimization of feature selection and model training. However, these methods are primarily limited to tabular datasets and do not account for cost awareness.

\textbf{Cost Aware Feature Selection}
Regarding comparisons to cost-aware FS methods such as CAFS \cite {momeni2021cafs} and others \cite{zhao2020cost}, these approaches typically focus on identifying a single global subset of features shared across all patients, assuming that all features are available upfront. In contrast, our method is specifically designed to reflect the realities of clinical practice, where patient information becomes available progressively, starting with basic, cost-free information and sequentially ordering additional tests based on observed findings, until a confident diagnosis is reached.

\section{Background on Robust Optimization:}
Unlike traditional optimization, which relies on precise knowledge of problem parameters, Robust Optimization (RO) addresses uncertainty by seeking solutions that perform well across a spectrum of potential scenarios. The goal of RO is to find a solution that minimizes the objective function under the worst-case realization of the uncertainty. In \cite{shaham2015understanding}, a minimization-maximization approach was introduced for training networks to optimize the following:
\begin{equation}
\min_{\theta} \tilde{J}(\theta, x, y) = \min_{\theta} \sum_{i=1}^{m} \max_{\tilde{x}_i \in \mathcal{U}_i} J(\theta, \tilde{x}_i, y_i), \notag 
\end{equation}
Where \( J(\theta, x, y) \) is the loss of a network with parameters \(\theta\) on \((x, y)\).
The optimization involves adjusting the network parameters \(\theta\) with respect to the worst-case data \(\tilde{x}_i\) from the uncertainty set \(\mathcal{U}_i\).
This is approximated through a two-step process. First, the network parameters \(\theta\) are held, an additive adversarial perturbation is calculated to each training example \(x_i\):

\begin{equation}
\Delta x_i = \arg \max_{\Delta:x_i+\Delta \in \mathcal{U}_i} J_{\theta, y_i} (x_i + \Delta) \notag
\end{equation}
When holding \(\theta\) and \(y\) fixed and viewing \( J(\theta, x, y) \) as a function of \(x\), we write \( J_{\theta, y}(x) \).
Subsequently, the network parameters $\theta$ are updated based on the perturbed data. However, finding the precise adversarial perturbation $\Delta x_i$ is usually not feasible, and full optimization is impractical. Therefore, in each iteration, a single ascent step approximates $\hat{\Delta}{x_i}$, followed by a single descent step to update $\theta$.

\section{The Proposed Approach}
In this section, we present our proposed method, P-CAFE, a novel model tailored for FS in EHR data. The primary objective is to predict patient outcomes, such as diagnosis, hospital readmission, treatment response, or mortality. The process begins with a patient's EHR, where the data is initially concealed. A representation of the patient is used, and in each iteration, a feature from the EHR is selected to test its value and update the representation. This online process, inspired by human-driven FS, continues until sufficient information is collected to predict the outcome.

\subsection{Rationale}
P-CAFE is specifically designed to address the challenges of EHR data and is characterized by several key attributes. First, it employs an online approach, progressively revealing one feature at a time in a human-like process, similar to how clinicians gather information step-by-step. 
Second, it is personalized, as the selection of each feature is guided by previously revealed features, tailoring the process to each individual patient. Third, it is multimodal, effectively integrating features from diverse data types within the EHR. Fourth, it is robust to the inherent sparsity often present in EHR data. Additionally, our method incorporates cost-aware feature selection, ensuring that the most informative subset of features is selected while adhering to a predefined budget. These attributes make our approach particularly well-suited for the unique complexities of EHR data.

\subsection{Problem Formulation}
Consider a dataset with \(n\) patients, each described by \(d\) features. For each patient \(i \in \{1, \ldots, n\}\), the feature set is represented as: $\{(f_1, f_2, \ldots, f_d) \mid f_j \in \mathcal{F}_j\}$. Here, \( f_j \) represents the value of the \( j \)-th feature. The feature space \(\mathcal{F}_j\) includes different types of data such as text, numeric values, and images. For example, \(\mathcal{F}_1\) represents the space of images, \(\mathcal{F}_2\) denotes the space of medical notes, and \(\mathcal{F}_3\) denotes lab tests. We introduce a cost vector \(C \in (\mathbb{R}^+)^d\)
, where \(C[j]\) represents the cost associated with acquiring feature \(j\). Additionally, we define a total budget \(B \in \mathbb{R}^+\), which constrains the cumulative cost of the selected features. 
Each patient's outcome, \( y_i \in \mathbb{R}^+ \), represents a supervised task, such as classification or regression. Examples of such tasks include predicting mortality or hospital readmission within a specified time window. Our objective is to identify a minimal, patient-specific subset of features that accurately predicts each patient's outcome \(y_i\). When feature costs are considered, the goal becomes selecting the most informative and predictive features within the given budget \(B\).

\subsection{MDP}  
We define the FS problem as an MDP, where an agent is responsible for selecting the optimal subset of features.  
The components of the MDP are defined as follows:

\textbf{State space}: Element-wise representation of the input and the mask: \[\{(f_1 m_1, f_2 m_2, \ldots, f_d m_d) \mid f_i \in \mathcal{F}, m_i \in \{0, 1\}\}\] The mask $m_i$ indicates whether a feature is revealed ($m_i = 1$) or not ($m_i = 0$). 

\textbf{Initial state}  
At the beginning of the episode, features available at no cost (e.g., age, gender) are marked as revealed (\(m_i = 1\)), while all other features are masked (\(m_i = 0\)).

\textbf{Action Space}: The action space is defined as \(\mathcal{A} = \{0, 1, \ldots, d\}\), where each action \( a \in \{0, 1, \ldots, d-1\} \) corresponds to revealing a specific feature, and the action \( a = d \) signifies the decision to predict the outcome. A key challenge when working with EHR data is the presence of patient-specific missing values, which arise because certain tests may not be applicable to all patients.  Our model is inherently robust to such sparsity, unlike traditional FS methods and classifiers that depend on complete feature availability and require explicit handling of missing values.
To address this issue, at the beginning of each episode, we ensure that unavailable features for a given patient have a zero probability of being selected. This adjustment guarantees that the action space remains valid and tailored to the patient's context.

\textbf{Transition rules}: The transition rules are deterministic.
    If the action \(a = k\) and \(k \in \{0, 1, \ldots, d-1\}\), the new state \(s'\) is obtained by taking the current state \(s\) and updating the feature mask \(m_k\) from \(0\) to \(1\).
Formally, if the current state $s = (f_1 m_1, \ldots, f_k \cdot 0, \ldots, f_d m_d)$  the new state $s' = (f_1 m_1, \ldots, f_k \cdot 1, \ldots, f_d m_d)$.
    If $a = d$, the state remains unchanged $s'=s$, and the current episode terminates \footnote{While this formulation describes the simplest case where each action reveals a single feature, our approach also supports scenarios where an action corresponds to a test that reveals multiple features simultaneously (e.g., a blood test that provides several lab values).}.

\textbf{Termination condition}: Either the agent chooses to make a guess $a=d$, or the episode reaches the pre-configured limit (number of steps, budget).

\textbf{Reward Function}: To guide the feature selection process, we introduce an additional model in our environment, referred to as the guesser. The guesser acts as the reward provider and is pre-trained as a supervised classifier. We defined two reward functions to guide the feature selection process: Gain-Based Reward and Guess-Based Reward. 

\begin{itemize}

\item Gain-Based Reward: When \( a < d \), the reward is determined by the increase in the probability mass assigned to the correct label. This is expressed as:  
\[
R(s', a) = \Pr(G(s') = y) - \Pr(G(s) = y),
\]
where \(s\) represents the current state, \(s'\) is the next state after taking action \(a\), \(G(s)\) denotes the guesser's prediction given the state \(s\), and \(y\) is the correct label.  

This formulation measures the improvement in confidence for the correct label as a result of selecting a new feature. By assigning rewards based on this gain, the approach emphasizes selecting features that contribute the most to refining the model's predictions, offering a systematic way to identify the most informative features.

If a cost vector \( C \) is provided, the reward is adjusted to account for the cost of the selected feature:  
\[
R(s', a) = \frac{\Pr(G(s') = y) - \Pr(G(s) = y)}{C[a]}.
\]  
This adjustment normalizes the reward by the cost, effectively measuring the gain per unit cost. More expensive features yield lower rewards unless they contribute significantly to the prediction accuracy.  

Additionally, as each feature is revealed, its cost is subtracted from the overall budget, ensuring that the model operates within resource constraints. This cost-sensitive adjustment encourages the selection of features that are both informative and cost-effective. Such an approach is particularly valuable in domains like healthcare, where resources are limited, and the trade-off between information gain and cost is critical.

\item Guess-Based Reward: When \( a = d \) (i.e., the process concludes), the reward is directly proportional to the probability of the guesser's prediction matching the actual label:  
\[
R(s, d) = \Pr(G(s) = y).
\]  
This reward reflects the model's confidence in its final prediction given the current state \( s \).
\end{itemize}

\subsubsection{The Agent (Feature Selector)} The agent reveals features at each step and updates its internal state with each new reveal. This process continues until the agent becomes sufficiently confident to predict the outcome, at which point it activates the guesser to make a prediction based on the revealed features.
As the reward function is designed to reflect the incremental value of each revealed feature, an agent that stops prematurely is unlikely to receive high rewards, as it may lack sufficient information. On the other hand, over-exploration is discouraged since, once additional features offer no predictive gain, the agent incurs extra cost without receiving any additional reward.

Our environment is designed to be compatible with any RL agent, allowing users the flexibility to choose and train their preferred RL agent. While the majority of the experimental results presented in this work are based on the Double Deep Q-Network (DDQN) agent \cite{van2016deep}, we also include experiments in Section ~\ref{sec:R_agents} that compare the performance of various RL agents.  
\footnote{The implementations of different RL agents were taken from the Stable-Baselines3 library \cite{JMLR:v22:20-1364}.}

\subsubsection{Agent and Guesser Training Dynamics}
\begin{figure}[h]
    \centering
    \includegraphics[width=0.5\textwidth]{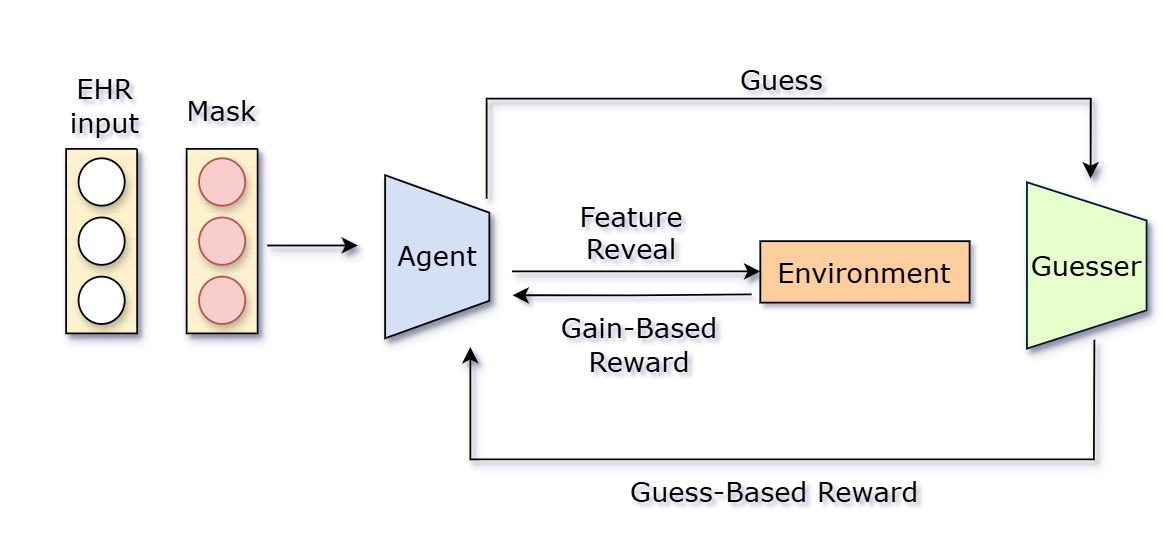}
\caption{The P-CAFE architecture. At each step, the agent reveals a feature and updates its internal state accordingly, receiving a Gain-Based reward. This process repeats until the agent attains sufficient confidence to predict the outcome, triggering the guesser to make a prediction based on the revealed features and receive a Guess-Based reward.}
    \label{fig:architecture}    
\end{figure}

The agent is responsible for selecting features to reveal, while the guesser uses these revealed features to predict the outcome and compute the agent's reward. The training process begins with the guesser, which is pre-trained as a supervised classifier. Once pre-training is complete, the agent is trained using the pre-trained guesser as a fixed model to guide its decisions. This process is illustrated in Figure \ref{fig:architecture}, which provides a clear overview of the architecture and its components.

\subsection{Handling Multimodal Data}
\subsubsection{Pretraining Phase}
The guesser is designed to effectively process multimodal data by generating an embedding vector for various input types. Each input type is embedded using a method tailored to its characteristics:

\begin{itemize}
    \item \textbf{Clinical Text Reports:} These are processed using \textit{Bio-ClinicalBERT}~\cite{alsentzer2019publicly}. A subsequent trained layer reduces the dimensionality of the text embeddings.
    \item \textbf{Image Data:} Embedded using the \textit{ResNet-50} model~\cite{he2016deep}, which provides a robust representation of the visual information.
    \item \textbf{Numeric Values:} Incorporated directly into the model without any additional modification, ensuring simplicity and efficiency.
    \item  \textbf{Time Series Data:} We use an LSTM (Long Short-Term Memory) network \cite{hochreiter1997long} to process the historical time steps in the input sequence, excluding the most recent time step. The LSTM captures temporal dependencies by generating hidden states at each time step. To create a fixed-size embedding for the sequence, we extract the hidden state from the final time step of the LSTM (which processes all but the most recent time step). We then concatenate the value of the most recent time step to this embedding. This approach places greater emphasis on the most recent time step while still preserving the embeddings of the entire history, ensuring that both the recent context and the long-term temporal dependencies are captured.
\end{itemize}

The guesser combines these diverse data types into a unified numeric vector, enabling it to train a network capable of integrating multimodal information.

\subsubsection{Agent Training} 
The agent manage a placeholder for each feature within the state vector. When a feature-reveal action is performed, the corresponding embedding is computed using the pre-trained embedding layers. This embedding is then placed in its designated position within the state vector, facilitating the integration of information from multiple data modalities.

\subsection{Avoiding Non-Stationarity through Robust Optimization}
In RL, the environment typically remains fixed throughout training, ensuring the stationarity of the Markov Decision Process (MDP). However, a non-stationary MDP arises when the environment changes during training. In our framework, the reward signal is derived from the guesser network. As the guesser requires training alongside the agent, this creates a feedback loop where both the agent and the guesser influence each other's training dynamics. While training both components together allows them to synchronize their learning, it introduces the risk of a non-stationary MDP, as changes in the guesser affect the reward function received by the agent.

To mitigate this issue, we employ robust optimization techniques to pre-train the guesser. By enhancing the guesser's ability to handle worst-case inputs, this approach eliminates the need for simultaneous training of the guesser and the agent. Consequently, the reward function remains stable, ensuring a fully stationary MDP and improving the overall training stability. The stationarity of MDP allows P-CAFE to support both on-policy and off-policy agents, by so significantly extending the approach introduced in \cite{shaham2020learning}. Our approach is divided into two sequential phases: robust pre-training of the guesser, followed by training of the agent.

During guesser pre-training, two masking strategies, adversarial and random masking, are employed to prepare the guesser for diverse input scenarios. Random masking is the primary strategy, ensuring alignment between pre-training and agent training distributions. Adversarial masking, on the other hand, is introduced with a low and gradually increasing probability to enhance robustness.

\textbf{Adversarial Masking:}
Robust Optimization (RO) \cite{ben2009robust} is utilized to enhance the model's robustness and improve generalization, as demonstrated in \cite{shaham2015understanding}, \cite{madry2017towards}. Inspired by this method, we designed an adversarial pre-training approach for the guesser. Using the idea explained in Section 3, we defined \(\mathcal{U}_i\) as a \( l_1 \) norm ball, which results in a sparse perturbation. In this perturbation, only one or a few of the entries of \(x_i\) are changed, specifically those with the largest magnitude in \(\nabla J_{\theta, y_i} (x_i)\). In our implementation, we create adversarial perturbations \(\hat{\Delta}_{x_i}\) by zeroing the features with the most significant gradients; those with the greatest influence on the input. 

The core idea of this approach is to challenge the guesser by concealing the features that contribute most significantly to reducing the loss function. By zeroing out these influential features, the guesser is forced to adapt and find alternative pathways to minimize the loss. This encourages the development of a more versatile and capable guesser, better equipped to generalize across various scenarios.

\textbf{Random Masking:} 
During agent training, the guesser processes samples where only the selected features are revealed, while the remaining features remain hidden. To ensure alignment between the input distributions of the pre-training and agent training phases, we incorporate random masking during pre-training. For each sample, a randomly chosen subset of features is concealed, and the guesser is trained on this masked dataset. This approach promotes consistency across training stages, minimizes the risk of distributional shifts, and enhances the overall performance and robustness of the model.

\subsection{Additional Design Choices For DDQN} 
\textbf{Priority Replay Memory:} Our method uses priority replay memory \cite{schaul2015prioritized}, to address the limitations of traditional uniform replay memory. In RL, replay memory is a repository of experiences used to train the DDQN. Conventional uniform replay memory treats all experiences equally, which can dilute the learning process by including less informative data.

In contrast, priority replay memory assigns greater importance to experiences with higher temporal-difference (TD) errors. This prioritization ensures that the agent focuses more on learning from significant experiences, thereby improving the efficiency of the learning process.

\textbf{Huber Loss:} Aimed at model robustness, we opted for Huber loss \cite{huber1992robust} as our preferred loss function, which has better resilience against outliers than MSE loss, that tends to penalize large errors disproportionately.

\section{Experiments}
Our experiments highlight that P-CAFE is highly suited for EHR datasets. On MIMIC-III, we showcase P-CAFE's ability to handle multi-modal data types, its robustness in the presence of inherently sparse data, support for personalized FS, and effective management of costs. On eICU, we illustrate P-CAFE's capability to handle time-series data. Clinical interpretability is shown in Supplementary Materials Section 9.5, where the selected features align with established medical knowledge.

\subsection{Electronic Health Record Datasets}
We have utilized the widely known MIMIC-III and eICU EHR databases for our experiments. Additionally, for further experiments on other EHR datasets, we refer the reader to the Supplementary Material. 
We emphasize that both MIMIC-III and eICU are widely recognized as standard benchmarks in EHR research. MIMIC-III, in particular, has been extensively used in the literature, with many studies relying on it exclusively. MIMIC-III and its variants incorporate multiple data modalities, while the eICU dataset includes rich time-series information. These datasets were chosen to enable meaningful comparisons with existing approaches and to align with canonical baselines in the field. Nonetheless, our framework is general and can be extended to other domains, including those involving rare diseases.
\subsubsection{MIMIC-III Datasets}
MIMIC-III is a widely recognized and publicly available dataset of EHR in raw format, serving as the primary benchmark in EHR studies. It has been utilized extensively in research studies, offering comprehensive information about patients admitted to critical care units at a large tertiary hospital \cite{johnson2016mimic}. Hosted on the PhysioNet repository \cite{mimiciii2015}, MIMIC-III provides an invaluable resource for advancing research and conducting detailed analyses in the field of healthcare. To benchmark critical aspects of working with EHR datasets, we used three distinct subsets of the MIMIC-III database:
\begin{itemize}
\item MIMIC-III Numeric: We utilized the data pipeline described in \cite{harutyunyan2019multitask}, a widely recognized benchmark for in-hospital mortality prediction based on the first 48 hours of an ICU stay. This dataset includes over 31 million clinical events encompassing 17 clinical variables, drawn from 42,276 ICU stays involving 33,798 unique patients.

\item MIMIC-III Costs: In the absence of publicly available EHR datasets that assign costs to each feature, we have enhanced the MIMIC-III Numeric dataset by assigning costs to individual features. These costs are based on the estimated effort in terms of time, money, and human resources required for each feature. These costs are fixed and dataset-specific. A detailed breakdown of these costs is provided in the Supplementary Materials.
\end{itemize}
The code to generate the datasets used in this study is available on \footnote{\url{https://github.com/shaham-lab/P-CAFE}}.

\subsubsection{eICU Collaborative Research Database}

The eICU Collaborative Research Database \cite{pollard2018eicu} is a large, publicly available multi-center dataset containing detailed clinical information from over 200,000 ICU stays across the United States. It was collected from hospitals using the eICU system developed by Philips Healthcare. A key feature of this dataset is its rich time-series structure, including vital signs, laboratory measurements, and treatment records recorded at regular intervals. For our experiments, we used the extraction pipeline introduced by \cite{sheikhalishahi2020benchmarking}, which focuses on in-hospital mortality prediction using the first 48 hours of ICU data. This pipeline generates time-series data that incorporates both continuous and categorical clinical features. The resulting dataset includes records from 24,629 patients, each with a variable number of time steps and 24 clinical features recorded per step.

\subsection {Evaluation Metrics}

\paragraph{Intersection Over Union (IoU):} 
IoU is calculated by dividing the size of the feature set common to all patients in the test set by the size of the combined feature set selected for all patients in the test set. A low IoU indicates that while the feature sets are diverse, there is minimal overlap between the features chosen for different patients. This supports a personalized approach tailored to each individual's unique needs, a lower IoU demonstrates a higher degree of personalization.

\paragraph{Cost:} The average cost of selected features per patient. There is a tradeoff between reducing the cost of features and gaining accurate results. This balance is especially critical in healthcare, as acquiring certain features can be costly in terms of time, resources, and patient discomfort. By reducing the cost healthcare providers can minimize unnecessary tests, leading to more efficient resource allocation.
While real-world constraints aren't strictly numerical, they exist in terms of time, cost, and effort. Using a budget during evaluation provides a practical way to assess how well the model balances accuracy and resource use. We refer the reader to Supplementary Section C.3 for an analysis of P-CAFE’s performance under varying budgets.

\paragraph{Accuracy:} Since discussing supervised learning, we have utilized accuracy as an evaluation metric.

\paragraph{AUC-ROC:}
The Area Under the Receiver Operating Characteristic Curve (AUC-ROC) is a key metric for evaluating model performance across all classification thresholds. It provides a balanced perspective on sensitivity and specificity, making it especially relevant in healthcare settings, where datasets are often imbalanced, and carefully managing the trade-off between false positives and false negatives is critical.

\paragraph{AUPRC:}
The Area Under the Precision-Recall Curve (AUPRC) offers deeper insights for highly imbalanced datasets by emphasizing precision and recall. This metric highlights the model's ability to effectively identify the minority class, making it an essential complement to AUC-ROC in healthcare applications.

\paragraph{Training Efficiency:} 
Although our approach relies on RL during training, the setup remains practical and scalable for real-world deployment. This is enabled by high data efficiency, achieved through the use of a strategic reward function provided by the guesser and short, patient-specific episodes. 

\subsection{Comparison to Baselines Using All Features}
In~\cite{harutyunyan2019multitask}, the authors introduced linear regression models alongside various neural network architectures, including experiments with a standard LSTM-based neural network and an enhanced variant, the channel-wise LSTM. Furthermore, they investigated both LSTM models in combination with a deep supervision approach, providing comprehensive insights into their performance. 
Below, we compare P-CAFE to all baseline methods reported in \cite{harutyunyan2019multitask}. As shown in Table~\ref{tab:performance_comparison}, P-CAFE achieves superior predictive performance compared to the baselines. Notably, P-CAFE is an FS method, so while all baseline methods utilized the full set of 17 features from the dataset with an IoU value of 1, P-CAFE demonstrated its effectiveness using only 13 features and achieved an IoU value of 0.29.

\begin{table}[H]
    \caption{Performance comparison of various models on the in-hospital mortality prediction task.}
    \centering
    \scriptsize 
    \renewcommand{\arraystretch}{1.2} 
    \setlength{\tabcolsep}{1mm} 
    \begin{tabular}{lrr}
        \toprule
        \textbf{Model} & \textbf{AUC-ROC} & \textbf{AUPRC} \\
        \midrule
        Logistic Regression                  & 0.848 (0.828, 0.868) & 0.474 (0.419, 0.529) \\ 
        Standard LSTM                        & 0.855 (0.835, 0.873) & 0.485 (0.431, 0.537) \\ 
        Standard LSTM + Deep Supervision     & 0.856 (0.836, 0.875) & 0.493 (0.438, 0.549) \\
        Channel-wise LSTM                    & 0.862 (0.844, 0.881) & 0.515 (0.464, 0.568) \\ 
        Channel-wise LSTM + Deep Supervision & 0.854 (0.834, 0.873) & 0.502 (0.447, 0.554) \\ 
        Multitask Standard LSTM             & 0.861 (0.842, 0.878) & 0.493 (0.439, 0.548) \\ 
        Multitask Channel-wise LSTM          & 0.870 (0.852, 0.887) & 0.533 (0.480, 0.584) \\ 
        P-CAFE                              & \textbf{0.901 (0.891, 0.921)} & \textbf{0.572 (0.551, 0.582)} \\
        \bottomrule
    \end{tabular}

    \label{tab:performance_comparison}
\end{table}

\subsection{Comparison Versus FS Methods}
We evaluate P-CAFE under varying cost budgets in comparison to other FS methods. Table \ref{tab:fs_models_comparison} illustrates that P-CAFE achieves the highest AUC-ROC and  AUPRC scores while maintaining low IoU and cost. A key advantage of our design is the ability to define the budget as a hyperparameter, enabling users to effectively balance predictive performance and cost constraints.
\begin{table}[H]
    \caption{Performance comparison of various FS models on the in-hospital mortality prediction task using the MIMIC-III dataset with costs.}
    \centering
    \scriptsize 
    \renewcommand{\arraystretch}{1.2} 
    \setlength{\tabcolsep}{1mm} 
    \begin{tabular}{lrrrr}
        \toprule
        \textbf{Model} & \textbf{AUC-ROC} & \textbf{AUPRC} & \textbf{Cost} & \textbf{IoU} \\
        \midrule
        Decision-Tree              & 0.620 (0.601, 0.632) & 0.204 (0.195, 0.210) & 40.00   & 0.81 \\
        XGBoost                    & 0.630 (0.612, 0.643) & 0.277 (0.252, 0.292) & 39.21   & --   \\
        IG-Random Forest           & 0.616 (0.602, 0.623) & 0.278 (0.254, 0.293) & 23.30   & 1.00 \\
        MI-Random Forest           & 0.606 (0.597, 0.612) & 0.264 (0.244, 0.283) & 25.10   & 1.00 \\
        RFE-Random Forest          & 0.601 (0.587, 0.622) & 0.262 (0.242, 0.288) & 32.20   & 1.00 \\
        LSPIN                      & 0.820 (0.812, 0.832) & 0.496 (0.482, 0.521) & 22.90   & 0.50 \\
        P-CAFE (Budget: 20)        & 0.880 (0.865, 0.901) & 0.530 (0.511, 0.553) & \textbf{16.18} & \textbf{0.25} \\
        P-CAFE (Budget: 30)        & \textbf{0.901 (0.891, 0.921)} & \textbf{0.572 (0.551, 0.582)} & 21.99   & 0.26 \\
        \bottomrule
    \end{tabular}
    \label{tab:fs_models_comparison}
\end{table}

\subsection{Comparison Versus a Personalized FS Method}
Existing personalized FS methods, such as the approach proposed by \cite{yang2022locally}, are not well-suited for EHR data due to their reliance on tabular datasets consisting exclusively of numeric features. These methods struggle to address the complexities of EHR data, which frequently include multimodal characteristics, time series, and inherently sparse data.

We compared our model to \cite{yang2022locally} to demonstrate its advantages over state-of-the-art generic personalized FS methods. In \cite{yang2022locally}, a regularization-based method was introduced for personalized FS. They introduce a probabilistic method that addresses the non-differentiability of the $\ell_0$ norm. Replacing binary indicators with Bernoulli vectors facilitates personalized FS that remains stable across similar samples. They design a locally sparse neural network where the personalized sparsity is learned to identify the subset of the most relevant features for each sample. This personalized sparsity is predicted via a gating network trained simultaneously with a prediction network and learns the most informative features for each sample. 

EHR data is inherently sparse due to the selective nature of medical testing. Missing values occur naturally because certain tests are only relevant or safe for specific patient populations and because patients typically undergo only a subset of the extensive array of available tests, based on their unique medical needs. The method proposed by \cite{yang2022locally} analyzes the entire feature space without accounting for patient-specific sparse data.

\begin{figure}[h]
    \centering
    \includegraphics[width=0.6\linewidth]{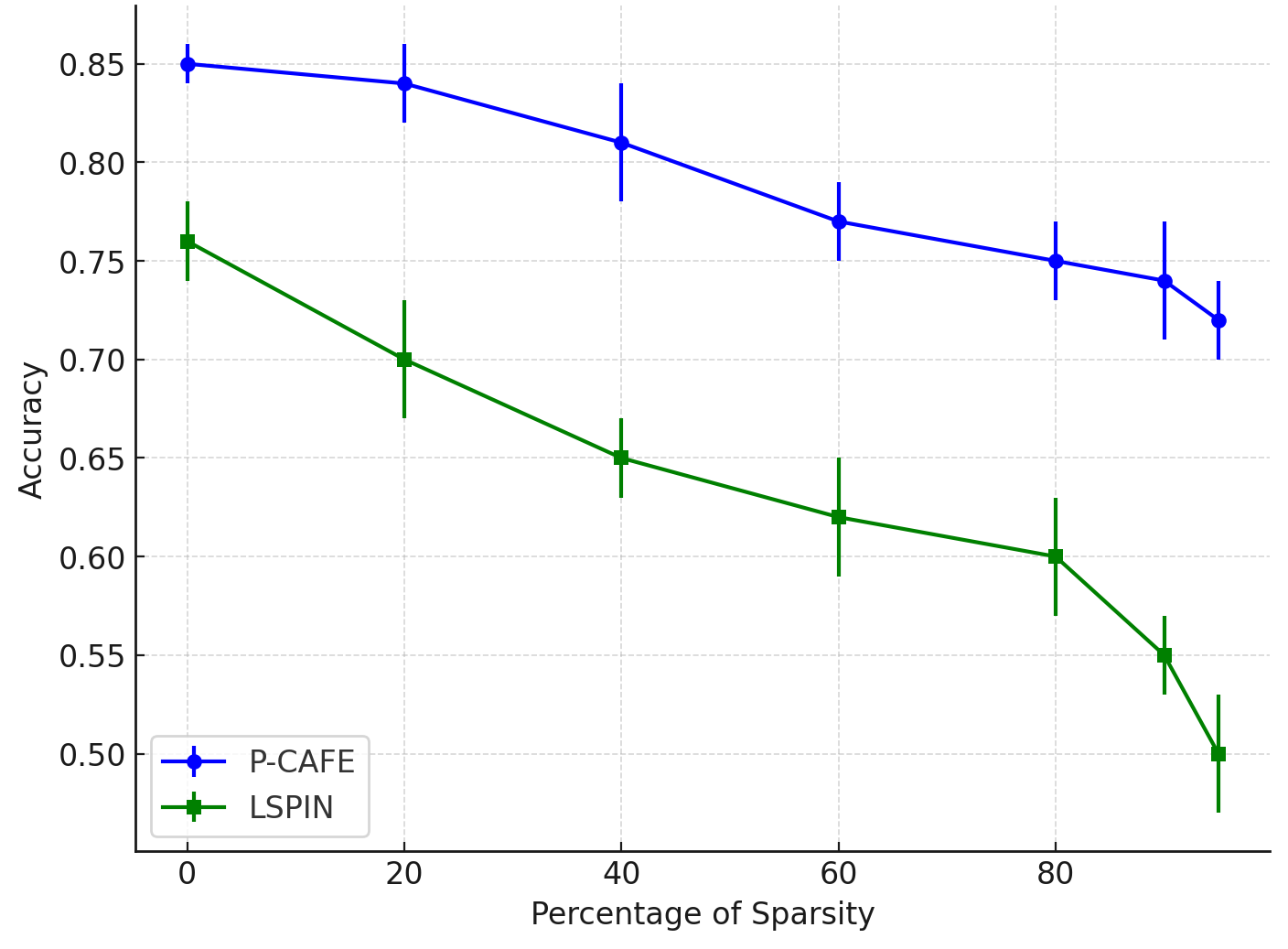} 
    \caption{Performance Comparison of P-CAFE and LSPIN }
    \label{fig:LSPIN}
\end{figure}

We utilized the MIMIC-III Numeric dataset and introduced varying levels of sparsity per patient to simulate real-world challenges. As shown in Figure \ref{fig:LSPIN}, P-CAFE exhibits a significant improvement in handling sparse data, achieving higher accuracy.
This performance can be attributed to P-CAFE adaptive action space where as previously explained at the beginning of each episode, we ensure that unavailable features for a given patient have a zero probability of being selected. This adjustment guarantees that the action space remains valid and tailored to the patient's context.

\subsection{Qualitative Example}
 As shown in Figure \ref{fig:cases}, P-CAFE selects features in stages, personalizing the FS process to individual cases. It integrates demographic, laboratory, categorical, and textual data.

\subsection{Ability to Handle Multi-Modal Data Types}
We evaluated P-CAFE using the MIMIC-III Multi-Modal dataset, which we partitioned into three subsets: numeric features only, textual features only, and multi-modal data (a combination of all features). As summarized in Table \ref{tab:Multi}, P-CAFE demonstrated the ability to effectively leverage information from various data modalities, achieving the highest performance when utilizing the multi-modal dataset. This highlights its capacity to integrate multi-modal data types.
We provide only our performance results due to the lack of available baselines that specifically address multimodal data types and feature selection while accounting for feature costs.

\begin{table}[H]
    \caption{Performance comparison of P-CAFE on different modalities of the MIMIC-III Multi Modal dataset}
    \centering
    \scriptsize 
    \renewcommand{\arraystretch}{1.2} 
    \setlength{\tabcolsep}{4mm} 
    \begin{tabular}{lrrr}
        \toprule
        \textbf{Dataset} & \textbf{Accuracy}  & \textbf{Cost} & \textbf{IoU}  \\        
        \midrule
         Numeric &  0.74 $\pm$0.02 & 16.03 $\pm$2.32 & 0.82 $\pm$0.02 \\
         Textual &  0.70 $\pm$0.03 & 4.02$\pm$1.02 & 0.93 $\pm$0.01 \\
         Multi-Modal & \textbf{0.83 $\pm$0.03} & \textbf{12.04$\pm$0.32} & \textbf{0.52 $\pm$0.02}\\
        \bottomrule
    \end{tabular}

    \label{tab:Multi}
\end{table}

\subsection {Ability to Handle Time-Series Data}
We utilize the eICU Database and extract the time-series data using the pipeline proposed in \cite{sheikhalishahi2020benchmarking}. A detailed explanation of our approach for handling time-series data is provided in Section 4.4.1. As shown in Table \ref{tab:eICU}, P-CAFE demonstrates superior predictive performance compared to the baselines presented in \cite{sheikhalishahi2020benchmarking}.
\begin{table}[H]
    \caption{Performance comparison of various models on the in-hospital mortality prediction task.}
    \centering
    \scriptsize 
    \renewcommand{\arraystretch}{1.2} 
    \setlength{\tabcolsep}{2mm} 
    \begin{tabular}{lrr}
        \toprule
        \textbf{Model} & \textbf{AUC-ROC} & \textbf{AUR-PR} \\
        \midrule
LR & 82.31 (81.56, 83.12) & 45.41 (44.01, 46.80) \\
ANN & 85.27 (84.69, 85.90) & 52.34 (51.01, 53.67) \\
BiLSTM & 86.55 (85.65, 87.52) & 54.98 (53.20, 56.77) \\
P-CAFE & \textbf{89.02 (87.66,89.02)} & \textbf{77.34 (76.23,78.43)} \\

        \bottomrule
    \end{tabular}

    \label{tab:eICU}
\end{table}

\subsection{Ablation Study}
We conduct an ablation study on the MIMIC-III Numeric dataset to evaluate the contribution of different components in the P-CAFE framework. As shown in Table~\ref{tab:ablation}, the complete P-CAFE model incorporating both the Agent and Guesser modules with robust training, achieves the best performance in terms of both AUC-ROC and AUPRC. This demonstrates the importance of each component in achieving optimal predictive performance.

\begin{table}[H]
    \caption{Ablation study on the MIMIC-III Numeric dataset.}
    \centering
    \scriptsize
    \renewcommand{\arraystretch}{1.2}
    \setlength{\tabcolsep}{1mm}
    \begin{tabular}{ccc|cc}
        \toprule
        \textbf{Agent} & \textbf{Guesser} & \textbf{Robust Training} & \textbf{AUC-ROC} & \textbf{AUPRC} \\
        \midrule
        
        $\times$ & $\checkmark$ & $\checkmark$ & 0.850 (0.842, 0.861) & 0.501 (0.492, 0.511) \\
        $\checkmark$ & $\checkmark$ & $\times$ & 0.877 (0.874, 0.878) & 0.550 (0.521, 0.573) \\
        $\checkmark$ & $\checkmark$ & $\checkmark$ & \textbf{0.901 (0.891, 0.921)} & \textbf{0.572 (0.551, 0.582)} \\
        \bottomrule
    \end{tabular}
    \label{tab:ablation}
\end{table}

\subsection{Compatibility With Various RL Agents}
\label{sec:R_agents}
Our environment is designed to support a wide range of RL agents, accommodating both on-policy and off-policy algorithms. To demonstrate this versatility, we evaluated several RL agents from the Stable-Baselines3 library \cite{JMLR:v22:20-1364} on the MIMIC-III Numeric dataset.

As shown in Table \ref{tab:Agents}, the best-performing agent for the MIMIC-III Numeric dataset is DQN. However, the choice of the best agent can vary depending on the specific characteristics of the dataset, allowing users to select the most appropriate agent for their needs.
\begin{table}[H]
    \caption{Performance of different RL agents on the MIMIC-III Numeric dataset.}
    \centering
    \scriptsize 
    \renewcommand{\arraystretch}{1.2} 
    \setlength{\tabcolsep}{3mm} 
    \begin{tabular}{lrrr}
        \toprule
        \textbf{RL Algorithm} & \textbf{Policy Type} & \textbf{Accuracy}  & \textbf{Features Count} \\        
        \midrule
        PPO & On-Policy & 0.767$\pm$0.02 & 11.150$\pm$0.04\\
        DQN & Off-Policy &  \textbf{0.772$\pm$0.03} & \textbf{10.73$\pm$0.05} \\
        A2C & On-Policy & 0.747$\pm$0.05  & 10.853$\pm$0.07\\
        \bottomrule
    \end{tabular}

    \label{tab:Agents}
\end{table}

\section{Conclusion}
This paper presents a novel, personalized cost-aware, online feature selection method tailored specifically for electronic health record datasets. Drawing inspiration from the online and personalized decision-making processes employed by healthcare professionals during clinical diagnosis, our approach addresses the inherent complexities of healthcare data. We developed a robust model capable of effectively handling multimodal data while integrating an advanced optimization technique for the guesser. A key innovation of our method is its cost-aware feature selection capability, which allows for identifying the optimal subset of features within a predefined budget, a critical consideration in resource-constrained healthcare environments. Additionally, our approach accommodates real-world healthcare scenarios by managing patient-specific sparsity.
Finally, this work also contributes a publicly available multimodal EHR dataset designed as a resource for the research community.

\bibliography{arxiv/P-CAFE}
\bibliographystyle{IEEEtranN}

\clearpage
\newpage
\section*{Supplementary Materials}

\section{MIMIC-III Multi-Modal Dataset:}
The dataset focuses on each patient's most recent ICU stay, ensuring it reflects the most relevant clinical context. For laboratory events, binary features were generated, where a value of '1' indicates an abnormal result, enabling quick identification of clinically significant values. Diagnoses were categorized into key groups based on typical conditions (e.g., cardiovascular or metabolic issues), with one-hot encoding creating binary indicators for each group. The dataset also incorporates clinical textual notes as distinct features, offering detailed insights into patient care. These notes span various categories, such as ECG, Echo, General, Nursing, Nutrition, Pharmacy, Physician, and Radiology, providing qualitative data to complement the numerical features.

\begin{table}[H]
    \caption{Feature Grouping For MIMIC-III Multi-Modal Dataset.}
     \scriptsize
     \centering
    \renewcommand{\arraystretch}{1.4} 
    \setlength{\tabcolsep}{1mm} 
    \begin{tabular}{p{1.5cm}p{6cm}}
        \toprule
        \textbf{Group} & \textbf{Features} \\
        \midrule
        Demographics & Gender, Age \\
        Lab Results & Base Excess, Glucose, Hemoglobin, Lactate Level, pCO$_2$, pH, pO$_2$, Albumin Level, Anion Gap, Bicarbonate Level, Total Bilirubin, Chloride Level, Potassium, Sodium Level, Hematocrit \\
        Diagnosis & Congestive Heart Failure, Cardiovascular Conditions, Gastrointestinal Conditions,Infectious Diseases, Metabolic Endocrine Disorders, Neurological Conditions, Renal Conditions, Respiratory Conditions, Trauma and Injury, Other Diagnoses \\
        Clinical Notes & Case Management Notes, Consultation Notes, Discharge Summary, ECG Report, Echocardiogram Report, General Notes, Nursing Notes, Nutrition Notes, Pharmacy Notes, Physician Notes, Radiology Report, Rehabilitation Services Notes, Respiratory Notes, Social Work Notes \\
        \bottomrule
    \end{tabular}
    \label{tab:feature_grouping}
\end{table}

\section{MIMIC-III With Costs:}
To simulate a realistic healthcare scenario, we have developed a cost framework for the features in this dataset. These costs reflect the estimated effort required for measurement and monitoring in a clinical setting, accounting for time, financial resources, and human effort.
Basic features, such as heart rate, temperature, and weight, are assigned a cost of 1, as they are routinely measured with minimal resource requirements. Complex features, such as the glasgow coma scale total, have a higher cost of 3, representing the effort involved in integrating multiple assessments. Advanced measurements, including fraction-inspired oxygen and oxygen saturation, are assigned a cost of 6, as they typically require specialized equipment or continuous monitoring.
The highest cost, 7, is attributed to laboratory-based tests like glucose and pH levels, which involve additional resources for processing.

\begin{table}[H]
    \caption{Feature Costs For MIMIC-III Dataset.}
    \centering
    \scriptsize 
    \renewcommand{\arraystretch}{1.2} 
    \setlength{\tabcolsep}{1mm} 
    \begin{tabular}{p{1cm}p{6cm}}
        \toprule
        \textbf{Cost} & \textbf{Features} \\        
        \midrule
        1 & Capillary refill rate, Glasgow coma scale eye opening, Glasgow coma scale motor response, Glasgow coma scale verbal response, Height, Temperature, Weight \\
        2 & Diastolic blood pressure, Heart Rate, Mean blood pressure, Respiratory rate, Systolic blood pressure \\
        3 & Glasgow coma scale total \\
        6 & Fraction inspired oxygen, Oxygen saturation \\
        7 & Glucose, pH level \\
        \bottomrule
    \end{tabular}

    \label{tab:feature_costs}
\end{table}

\section {Additional Experiments}
\subsection{Comparison Versus Global FS Methods}
This study highlights the advantages of personalized FS compared to global FS methods. We evaluated our P-CAFE approach against traditional global FS methods presented by \citep{zuo2021curvature} using the Diabetic Retinopathy Debrecen dataset \citep{Retinopathy}, which contains 1,151 instances with 19 features.

\begin{figure}[h]
    \centering
    \includegraphics[width=0.5\linewidth]{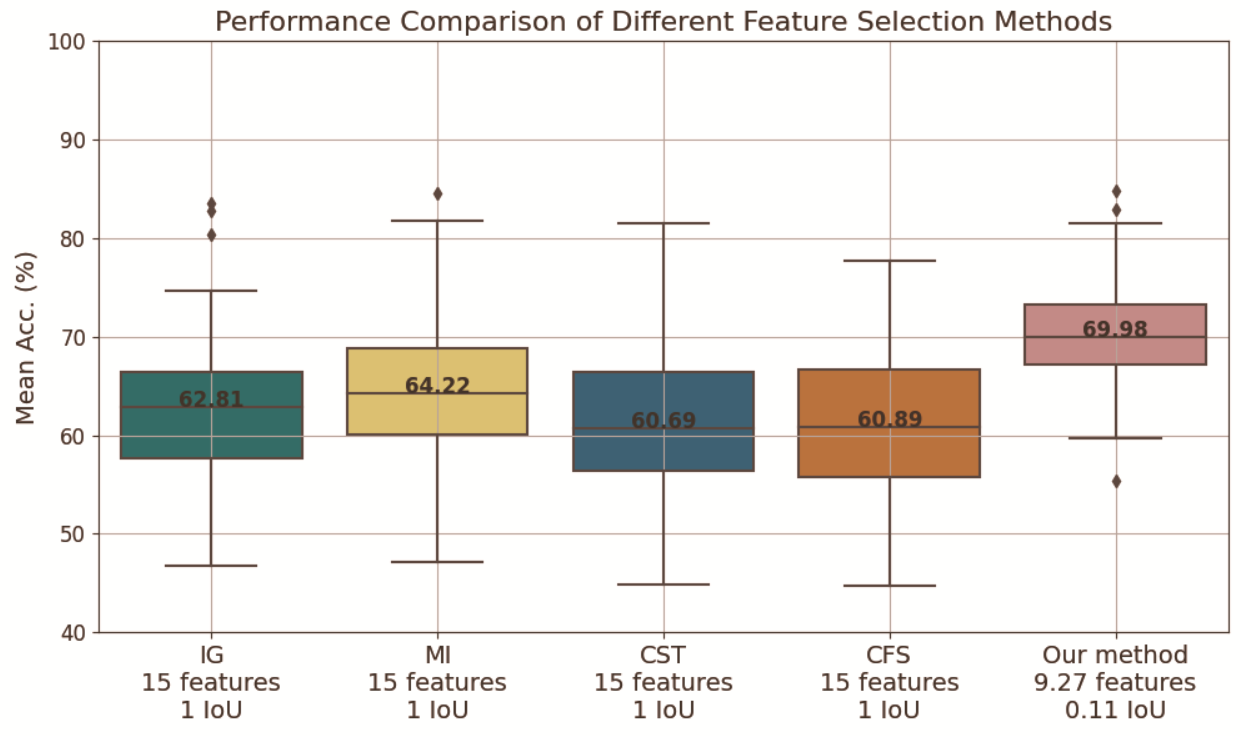} 
    \caption{Performance comparison of P-CAFE against CFS, IG, MI, and CST methods, as reported in \citep{zuo2021curvature}.}
    \label{fig:comparison}
\end{figure}

As shown in Figure \ref{fig:comparison}, our P-CAFE approach outperformed global FS methods by achieving higher accuracy and lower IoU while using fewer features. Since no code was available to apply the global FS methods to a new dataset, we utilized the dataset provided in the original paper to demonstrate our performance.

\subsection{Ability to Manage Cost Budget}
To evaluate our method's capability to incorporate feature costs and select the optimal feature subset within a given budget, we evaluated the performance of P-CAFE under different budgets. As shown in Table \ref {tab:P-CAFE_budget_accuracy}, allocating a higher budget enables the selection of more informative features, resulting in improved accuracy, until a limit is reached that adding a budget will not help in improvement. This demonstrates P-CAFE's effectiveness in balancing feature costs with predictive performance.

\begin{table}[H]
    \caption{Performance of P-CAFE under varying budgets on the MIMIC-III Costs dataset.}
    \centering
    \scriptsize 
    \renewcommand{\arraystretch}{1.2} 
    \setlength{\tabcolsep}{6mm} 
    \begin{tabular}{lr}
        \toprule
        \textbf{Budget} & \textbf{Accuracy} \\        
        \midrule
        10 & 0.783 \\
        20 & 0.801 \\
        30 & 0.821 \\
        40 & 0.812 \\
        50 & 0.823 \\
        \bottomrule
    \end{tabular}

    \label{tab:P-CAFE_budget_accuracy}
\end{table}

\subsection{Comparing To a RL-based Method}
This section will compare P-CAFE with an RL-based method for global FS proposed by \citep{rasoul2021feature}. Table 4 highlights the accuracy improvements for both datasets that were examined.

\begin{table}[H] 
    \caption{Comparison of P-CAFE with other RL approaches.}
    \centering
    \scriptsize 
    \renewcommand{\arraystretch}{1.2} 
    \setlength{\tabcolsep}{1mm} 
    \begin{tabular}{lcc}
        \toprule
        & \textbf{Connectionist Bench} \citep{Dataset} & \textbf{WPBC} \citep{BreastCancer} \\ 
        \midrule
        \citep{rasoul2021feature} Acc & 0.73 $\pm$ 0.108 & 0.76 $\pm$ 0.007 \\ 
        P-CAFE Acc                  & \textbf{0.82 $\pm$ 0.02} & \textbf{0.79 $\pm$ 0.01} \\
        \bottomrule
    \end{tabular}

    \label{tab:comparison_rl}
\end{table}

\subsection{Comparison to Shaham et al. (2020)}

Our objective is to develop a human-like FS framework specifically tailored to EHRs. While P-CAFE is inspired by  \citep{shaham2020learning} approach, their method is not applicable to EHR data since it is not able to handle multimodal data types and does not account for varying feature costs. Additionally,  \citep{shaham2020learning} highlight a challenge inherent in their framework, the non-stationarity of the MDP caused by the need to train the guesser model alongside the agent. They partially address this issue through an alternating training procedure, which provides local stability but fails to ensure global stability. In contrast, P-CAFE introduces a robust optimization technique (Section 4.5) that fully stabilizes the MDP, enabling more reliable and consistent learning. P-CAFE also employs a gain-based reward function based on the improvement in the guesser’s prediction confidence normalized by feature cost, encouraging strategic and cost-aware selection (Section 4.2). While the reward function used in \citep{shaham2020learning} is uniform and not related to the informativeness of the characteristics, leading to inefficient learning. Additional enhancements, such as replacing MSE with Huber loss and using prioritized over naive experience replay, further boost performance. Table \ref{tab:shaham} reports results on purely numerical data, as  \citep{shaham2020learning} only support this modality. As shown in Table \ref{tab:shaham}, our P-CAFE design outperforms the method proposed by \citep{shaham2020learning} across all metrics.

\begin{table}[H]
    \caption{Performance comparison on the Diabetes dataset \citep{Diabetes}. Results are averaged over 10 iterations.}
    \centering
    \scriptsize 
    \renewcommand{\arraystretch}{1.2} 
    \setlength{\tabcolsep}{0.8mm} 
    \begin{tabular}{lrrrrr}
        \toprule
        \textbf{Type} & \textbf{Acc} & \textbf{Features Count} & \textbf{Intersection Count} & \textbf{Union Count} & \textbf{IoU} \\        
        \midrule
        Shaham et al. (2020) & 0.69 $\pm$ 0.007 & 4 & 3 & 6 & 0.50 \\
        P-CAFE                    & \textbf{0.83 $\pm$ 0.01} & \textbf{2} & \textbf{1} & \textbf{6} & \textbf{0.16} \\
        \bottomrule
    \end{tabular}

    \label{tab:shaham}
\end{table}

\subsection{Clinical Interpretability}
\begin{figure}[h]
    \centering
    \includegraphics[width=0.8\textwidth]{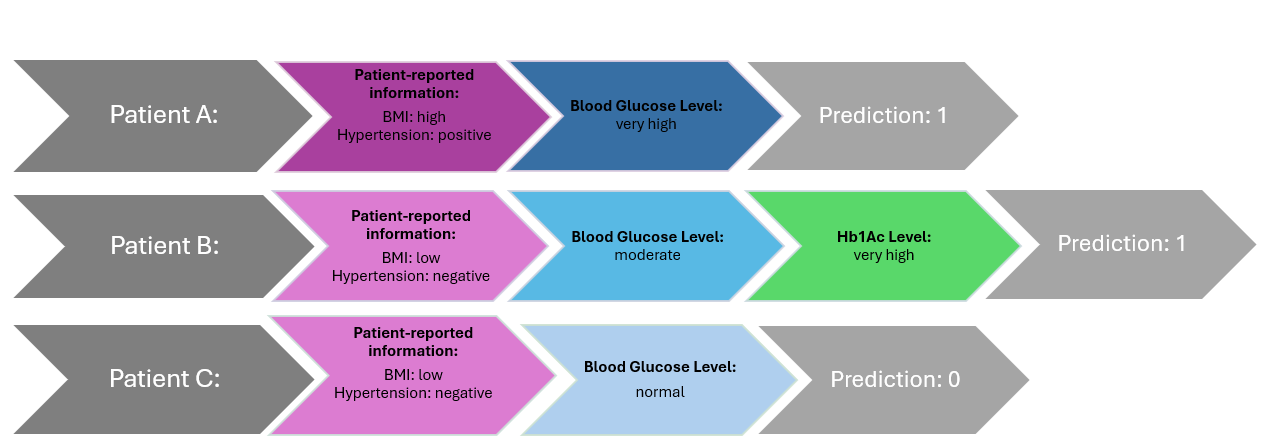}
\caption{Clinically Interpretable Feature Acquisition in Diabetes Diagnosis}
    \label{fig:Diabetes}    
\end{figure}

We conducted an additional experiment on the Diabetes Prediction dataset \citep{Diabetes}. The features used include gender, age, hypertension, heart disease, smoking history, and BMI. These are considered cost-free, as they rely on patient-reported information and do not require any medical tests. The blood glucose level feature, which reflects current blood sugar levels, requires a standard blood test and is assigned a cost of 1. In contrast, the HbA1c level feature provides an estimate of average blood sugar over the past 3 months and requires a more specific laboratory analysis, we assign it a higher cost of 3.
Figure \ref{fig:Diabetes} shows that for patients with a high blood glucose level and patient-reported information indicating poor health (e.g., high BMI and positive hypertension status, as in Patient A), the model confidently stops and predicts the patient as diabetic. In contrast, when the blood glucose level is moderate (Patient B), the model continues to acquire additional features (HbA1c level) before making a prediction, reflecting the need for further confirmation. For patients whose reported information indicates good health and who also exhibit normal glucose levels (Patient C), the model predicts a non-diabetic outcome without requesting further tests. This behavior demonstrates the model’s ability to adaptively halt costly testing when sufficient evidence has already been gathered.

\end{document}